\documentclass[11pt,a4paper]{article}
\usepackage[hyperref]{emnlp-ijcnlp-2019}
\usepackage{times}
\usepackage{latexsym}
\usepackage{float}
\usepackage{tabularx}
\usepackage{multirow}
\usepackage{url}
\usepackage{times}
\usepackage{latexsym}
\usepackage{amsmath}
\usepackage{amssymb}
\usepackage{tabularx}
\usepackage{url}
\usepackage{booktabs}
\usepackage{graphicx}
\usepackage{multirow}
\usepackage{todonotes}
\usepackage{array}
\usepackage{tikz}

\aclfinalcopy 
\title{Effective Use of Transformer Networks for Entity Tracking}

\author{Aditya Gupta \and Greg Durrett \\
  Department of Computer Science \\
  The University of Texas at Austin \\
  {\tt \{agupta,gdurrett\}@cs.utexas.edu} }

\date{}

\begin{document}
\maketitle
\begin{abstract}
Tracking entities in procedural language requires understanding the transformations arising from actions on entities as well as those entities' interactions. While self-attention-based pre-trained language encoders like GPT and BERT  have been successfully applied across a range of natural language understanding tasks, their ability to handle the nuances of procedural texts is still untested.  In this paper, we explore the use of pre-trained transformer networks for entity tracking tasks in procedural text. First, we test standard lightweight approaches for prediction with pre-trained transformers, and find that these approaches underperform even simple baselines. We show that much stronger results can be attained by restructuring the input to guide the transformer model to focus on a particular entity. Second, we assess the degree to which transformer networks capture the process dynamics, investigating such factors as merged entities and oblique entity references. On two different tasks, ingredient detection in recipes and QA over scientific processes, we achieve state-of-the-art results, but our models still largely attend to shallow context clues and do not form complex representations of intermediate entity or process state.\footnote{Code to reproduce experiments in this paper is available at \url{https://github.com/aditya2211/transformer-entity-tracking}}
\end{abstract}

\section{Introduction}


Transformer based pre-trained language models \cite{devlin2018bert, radford2018improving, radford2019language, m2019spanbert, yang2019xlnet}
have been shown to perform remarkably well on a range of tasks, including entity-related tasks like coreference resolution \cite{kantor-globerson-2019-coreference} and named entity recognition \cite{devlin2018bert}. This performance has been generally attributed to the robust transfer of lexical semantics to downstream tasks. However, these models are still better at capturing syntax than they are at more entity-focused aspects like coreference \cite{tenney-etal-2019-bert, tenney2018what}; moreover, existing state-of-the-art architectures for such tasks often perform well looking at only local entity mentions \cite{wiseman-etal-2016-learning, lee-etal-2017-end, peters-etal-2017-semi}  rather than forming truly global entity representations \cite{RahmanNg2009,lee-etal-2018-higher}. Thus, performance on these tasks does not form sufficient evidence that these representations strongly capture entity semantics. Better understanding the models' capabilities requires testing them in domains involving complex entity interactions over longer texts. One such domain is that of procedural language, which is strongly focused on tracking the entities involved and their interactions \cite{MoriEtAl2014,dalvi2018tracking, bosselut2018simulating}. 

\begin{figure*}[t!]
  \includegraphics[width=\textwidth]{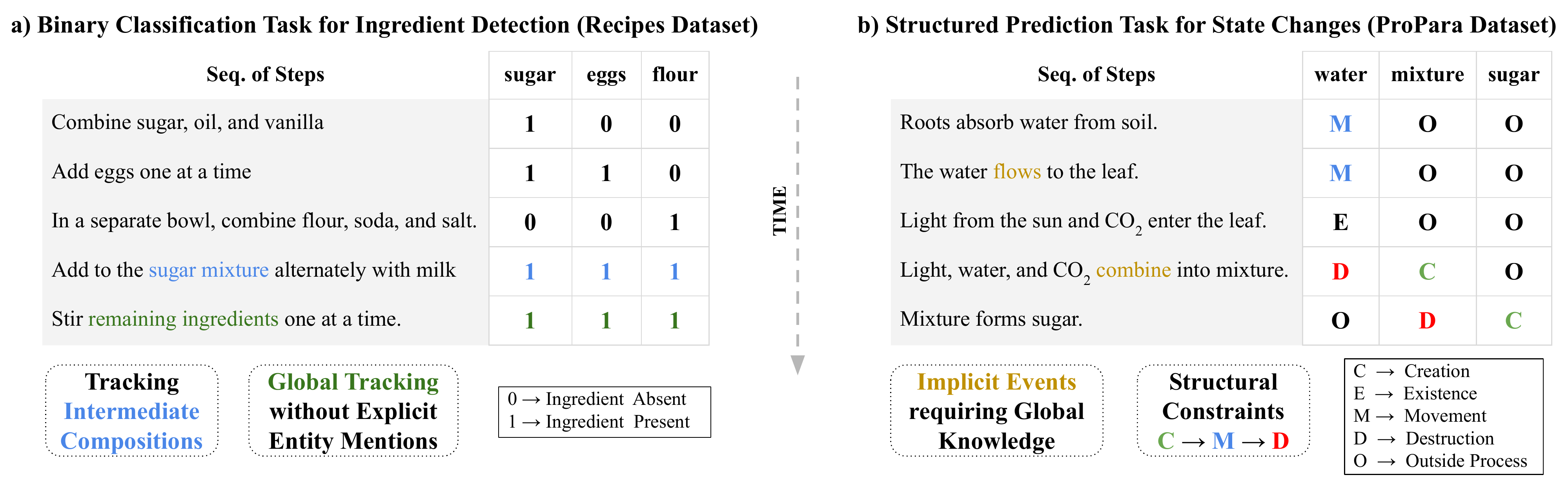}
  \centering
  \caption{Process Examples from (a) \textsc{Recipes} as a binary classification task of ingredient detection, and (b) \textsc{ProPara} as a structured prediction task of identifying state change sequences. Both require cross-sentence reasoning, such as knowing what components are in a \emph{mixture} and understanding verb semantics like \emph{combine}.}
  \label{fig:examples}
\end{figure*}

This paper investigates the question of how transformer-based models form entity representations and what these representations capture. We expect that after fine-tuning on a target task, a transformer's output representations should somehow capture relevant entity properties, in the sense that these properties can be extracted by shallow classification either from entity tokens or from marker tokens. However, we observe that such ``post-conditioning'' approaches don't perform significantly better than rule-based baselines on the tasks we study. We address this by proposing entity-centric ways of structuring input to the transformer networks, using the entity to guide the intrinsic self-attention and form entity-centric representations for all the tokens. We find that our proposed methods lead to a significant improvement in performance over baselines. 

Although our entity-specific application of transformers is more effective at the entity tracking tasks we study, we perform additional analysis and find that these tasks still do not encourage transformers to form truly deep entity representations. Our performance gain is largely from better understanding of verb semantics in terms of associating process actions with entity the paragraph is conditioned on. The model also does not specialize in ``tracking'' composed entities per se, again using surface clues like verbs to identify the components involved in a new composition.

We evaluate our models on two datasets specifically designed to invoke procedural understanding: (i) \textsc{Recipes} \cite{kiddon-etal-2016-globally}, and (ii) \textsc{ProPara} \cite{dalvi2018tracking}. For the \textsc{Recipes} dataset, we classify whether an ingredient was affected in a certain step, which requires understanding when ingredients are combined or the focus of the recipe shifts away from them. The \textsc{ProPara} dataset involves answering a more complex set of questions about physical state changes of components in scientific processes. To handle this more structured setting, our transformer produces potentials consumed by a conditional random field which predicts entity states over time. Using a unidirectional GPT-based architecture, we achieve state-of-the-art results on both the datasets; nevertheless, analysis shows that our approach still falls short of capturing the full space of entity interactions.

\section{Background: Process Understanding}
\label{sec:background}

Procedural text is a domain of text involved with understanding some kind of process, such as a phenomenon arising in nature or a set of instructions to perform a task. Entity tracking is a core component of understanding such texts. 

\citet{dalvi2018tracking} introduced the \textbf{\textsc{ProPara}} dataset to probe understanding of scientific processes. The goal is to track the sequence of physical state changes (creation, destruction, and movement) entites undergo over long sequences of process steps. Past work involves both modeling entities across time \cite{das2018building} and capturing structural constraints inherent in the processes \cite{tandon2018reasoning,gupta2019tracking} 
Figure~\ref{fig:examples}b shows an example of the dataset posed as a structured prediction task, as in \cite{gupta2019tracking}. For such a domain, it is crucial to capture implicit event occurrences beyond explicit entity mentions. For example, in \emph{fuel goes into the generator. The generator converts mechanical energy into electrical energy}'', the \emph{fuel} is implicitly destroyed in the process. 

\citet{bosselut2018simulating} introduced the task of detecting state changes in recipes in the \textbf{\textsc{Recipes}} dataset and proposed an entity-centric memory network neural architecture for simulating action dynamics.
Figure~\ref{fig:examples}a shows an example from the \textsc{Recipes} dataset with a grid showing ingredient presence. We focus specifically on this core problem of ingredient detection; while only one of the sub-tasks associated with their dataset, it reflects some complex semantics involving understanding the current state of the recipe. Tracking of ingredients in the cooking domain is challenging owing to the compositional nature of recipes whereby ingredients mix together and are aliased as intermediate compositions.

We pose both of these procedural understanding tasks as classification problems, predicting the state of the entity at each timestep from a set of pre-defined classes. In Figure~\ref{fig:examples}, these classes correspond to either the presence (1) or absence (0) or the sequence of state changes create (C), move (M), destroy (D), exists (E), and none (O).

State-of-the-art approaches on these tasks are inherently entity-centric. Separately, it has been shown that entity-centric language modeling in a continuous framework can lead to better performance for LM related tasks \cite{N18-1204, D17-1195}. Moreover, external data has shown to be useful for modeling process understanding tasks in prior work \cite{tandon2018reasoning,bosselut2018simulating}, suggesting that pre-trained models may be effective.



With such tasks in place, a strong model will ideally learn to form robust entity-centric representation at each time step instead of solely relying on extracting information from the local entity mentions. This expectation is primarily due to the evolving nature of the process domain where entities undergo complex interactions, form intermediate compositions, and are often accompanied by implicit state changes. We now investigate to what extent this is true in a standard application of transformer models to this problem.

\section{Studying Basic Transformer Representations for Entity Tracking}

\label{sec:ettn}

\subsection{Post-conditioning Models}

The most natural way to use the pre-trained transformer architectures for the entity tracking tasks is to simply encode the text sequence and then attempt to ``read off'' entity states from the contextual transformer representation. We call this approach \emph{post-conditioning}: the transformer runs with no knowledge of which entity or entities we are going to make predictions on, but we only condition on the target entity after the transformer stage.

Figure~\ref{fig:temps} depicts this model. Formally, for a labelled pair $(\{s_1, s_2, \dots, s_t\}, y_{et})$, we encode the tokenized sequence of steps up to the current timestep (the sentences are separated by using a special \texttt{[SEP]} token), independent of the entity. We denote by $X=[h_{1}, h_{2},\dots, h_{m}]$ the contextualized hidden representation of the $m$  input tokens from the last layer, and by $\textstyle  g_{e}\!=\!\!\!\sum\limits_{\text{ent toks}}\!emb(e_i)$ the entity representation for post conditioning. We now use one of the following two ways to make an entity-specific prediction:
\begin{figure}[t!]
  \includegraphics[width=\linewidth]{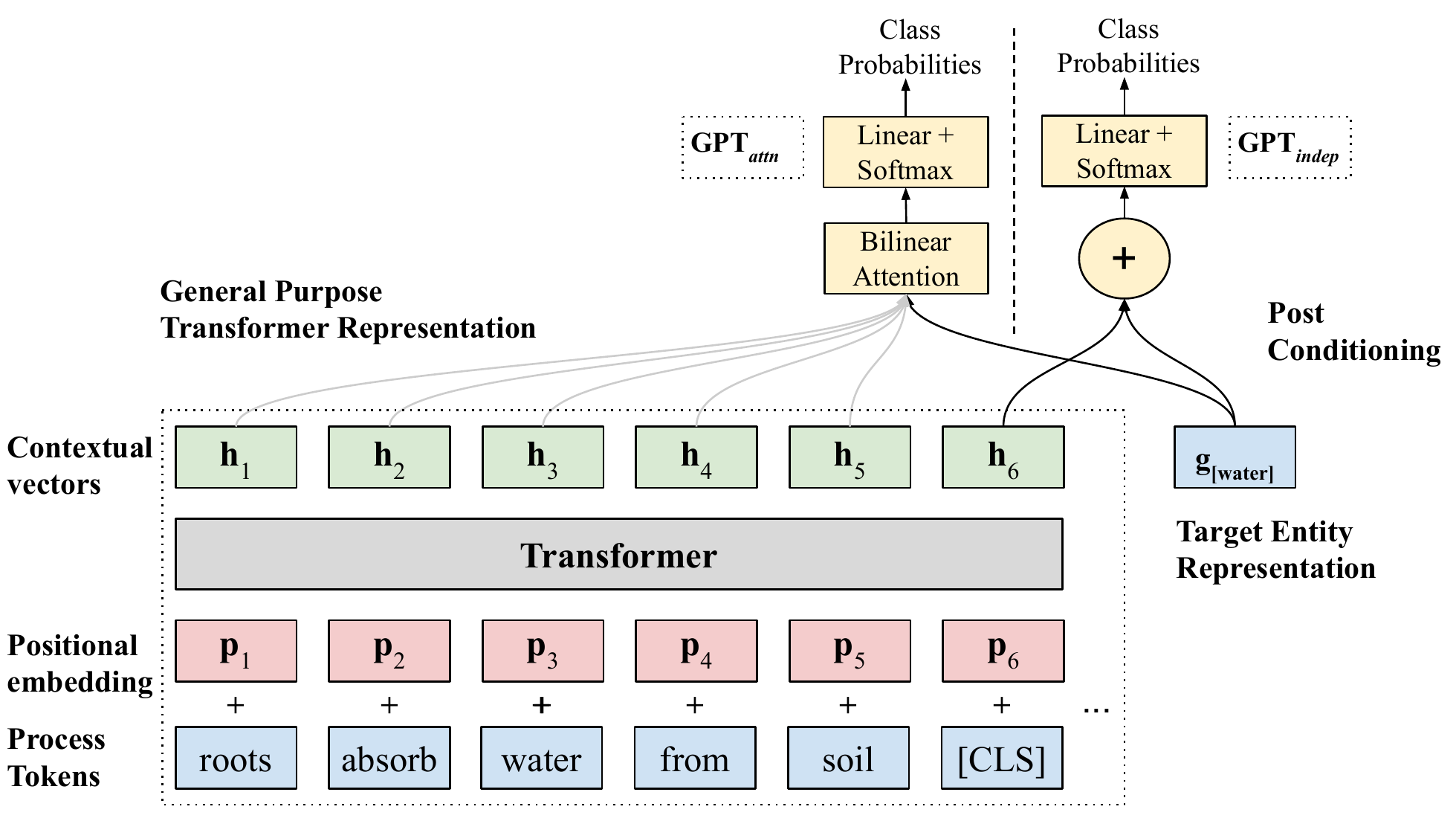}
  \centering
  \caption{Post-conditioning entity tracking models. Bottom: the process paragraph is encoded in an entity-independent manner with transformer network and a separate entity representation $g_{[water]}$ for post-conditioning. Top: the two variants for the conditioning: (i) GPT$_{attn}$, and (ii) GPT$_{indep}$.}
  \label{fig:temps}
\end{figure}
\paragraph{Task Specific Input Token} We append a $\texttt{[CLS]}$ token to the input sequence and use the output representation of the $\texttt{[CLS]}$ token denoted by $h_{ \texttt{[CLS]}}$ concatenated with the learned BPE embeddings of the entity as the representation $c_{e,t}$ for our entity tracking system. We then use a linear layer over it to get class probabilities:
\begin{equation*}
\begin{split}
   c_{e,t} &= [h_{\texttt{[CLS]}};g_e] \\
  P(y_t|s_t, s_{t-1}, &\dots, s_1, e) = \mathrm{softmax}( c_{e,t}W_{task})
\end{split}
\end{equation*}
The aim of the \texttt{[CLS]} token is to encode information related to general entity related semantics participating in the recipe (\emph{sentence priors}). We then use a single linear layer to learn sentence priors and entity priors independently, without strong interaction. We call this model 
GPT$_{indep}$.

\paragraph{Entity Based Attention} Second, we explore a more fine-grained way of using the GPT model outputs. Specifically, we use bilinear attention between $g_e$ and the transformer output for the process tokens $X$ to get a contextual representation $c_{e,t}$ for a given entity. Finally, using a feed-forward network followed by softmax layer gives us the class probabilities:
\vspace{-1em}
\begin{equation*}
\begin{split}
   a_i &= g_e^{T}* W_{sim}*h_i \\
   \alpha &= \text{softmax}(a) \\
   c_{e,t} &= \sum \alpha_i*h_i 
 \end{split}
\end{equation*}  
\begin{equation*}
\begin{split}
P(y_t|s_t, s_{t-1}, \dots, s_1, e) = \mathrm{softmax}( c_{e,t}W_{task})
 \end{split}
\end{equation*} 

\begin{table*}[t!]
\setlength{\tabcolsep}{3.5pt}
\small
\centering
\begin{tabular}{c|c}
\toprule
  \bf Variant & \bf Template \\

     \midrule
     Sentence, Entity First & \texttt{[START]} \textbf{Target Entity} \texttt{[SEP]} Steps $1$ to $t-1$ \texttt{[SEP]} Step $t$ \texttt{[CLS]}  \\
     \midrule
          Sentence, Entity Last & \texttt{[START]}Steps $1$ to $t-1$ \texttt{[SEP]} Step $t$ \texttt{[SEP]} \textbf{Target Entity} \texttt{[CLS]} \\
          \midrule
    Document, Entity First & \texttt{[START]} \textbf{Target Entity} \texttt{[SEP]} Step $1$ \texttt{[CLS]} Step $2$  \texttt{[CLS]} $\dots$ Step $T$  \texttt{[CLS]}  \\
     \midrule
    Document, Entity Last & \texttt{[START]} Step $1$ \texttt{[SEP]} \textbf{Target Entity} \texttt{[CLS]} $\dots$  Step $T$ \texttt{[SEP]} \textbf{Target Entity} \texttt{[CLS]}  \\   
  \bottomrule
 \end{tabular}
 \label{tab:variants_temps}
 \caption{Templates for different proposed entity-centric modes of structuring input to the transformer networks.}
\end{table*}

The bilinear attention over the contextual representations of the process tokens allows the model to fetch token content relevant to that particular entity. We call this  model GPT$_{attn}$. 

\subsection{Results and Observations}

We evaluate the discussed post-conditioning models on the ingredient detection task of the \textsc{Recipes} dataset.\footnote{We discuss training details more in Section~\ref{sec:training_details}, but largely use a standard GPT training protocol \cite{radford2018improving}.} To benchmark the performance, we compare to three rule-based baselines. This includes (i) \emph{Majority Class}, (ii) \emph{Exact Match} of an ingredient $e$ in recipe step $s_t$, and  (iii) \emph{First Occurrence}, where we predict the ingredient to be present in all steps following the first exact match. These latter two baselines capture natural modes of reasoning about the dataset: an ingredient is used when it is directly mentioned, or it is used in every step after it is mentioned, reflecting the assumption that a recipe is about incrementally adding ingredients to an ever-growing mixture. We also construct a LSTM baseline to evaluate the performance of ELMo embeddings (ELMo$_{token}$ and ELMo$_{sent}$) \cite{peters-etal-2018-deep} compared to GPT. 



Table \ref{tab:pc_models} compares the performance of the discussed models against the baselines, evaluating per-step entity prediction performance.  Using the ground truth about ingredient's state, we also report the uncombined (UR) and combined (CR) recalls, which are per-timestep ingredient recall distinguished by whether the ingredient is explicitly mentioned (uncombined) or part of a mixture (combined). Note that \emph{Exact Match} and \emph{First Occ} baselines represent high-precision and high-recall regimes for this task, respectively. 

\setlength{\tabcolsep}{4pt}
\begin{table}[h]
\small
\centering
\begin{tabular}{l|cccc|cc}
\toprule
  \bf Model & \bf P & \bf R & \bf $F_1$ & \bf Acc & \bf UR & \bf CR \\
  \midrule
   \multicolumn{7}{c}{ \small Performance Benchmarks} \\
  \midrule
  Majority & - & - & - & 57.27 & - & -\\
  Exact Match & \bf 84.94 & 20.25 & 32.70 & 64.39& 73.42 & 4.02 \\
  First Occ &  65.23  & \bf 87.17 & \bf{74.60} & \bf 74.65 & \bf 84.88 & \bf 87.79 \\
  \midrule
   \multicolumn{7}{c}{ \small Models} \\
  \midrule
    GPT$_{attn}$ & 63.94  & 71.72  & 67.60  & 70.63  & 54.30 & 77.04 \\
   GPT$_{indep}$ & 67.05  & 69.07 & 68.04 & 72.28 & 47.09 & 75.79 \\
    ELMo$_{token}$ & 64.96  &  76.64 & 70.32  & 72.35  & 69.14  & 78.94  \\
      ELMo$_{sent}$ & 69.09  & 72.88  & 70.90  & 74.48  & 57.05 & 77.71  \\
  \bottomrule
 \end{tabular}
 \caption{Performance of the rule-based baselines and the post conditioned models on the ingredient detection task of the \textsc{Recipes} dataset. These models all underperform \emph{First Occ}.}
 \label{tab:pc_models} 
\end{table}

As observed from the results, the post-conditioning frameworks underperform compared to the \emph{First Occ} baseline. While the CR values appear to be high, which would suggest that the model is capturing the addition of ingredients to the mixture, we note that this value is also lower than the corresponding value for \emph{First Occ}. This result suggests that the model may be approximating the behavior of this baseline, but doing so poorly. The unconditional self-attention mechanism of the transformers does not seem sufficient to capture the entity details at each time step beyond simple presence or absence. Moreover, we see that GPT$_{indep}$ performs somewhat comparably to GPT$_{attn}$, suggesting that consuming the transformer's output with simple attention is not able to really extract the right entity representation.

For \textsc{ProPara}, we observe similar performance trends where the post-conditioning model performed below par with the state-of-the-art architectures.
 

\section{Entity-Conditioned Models}
\label{sec:ec_models}
The post-conditioning framework assumes that the transformer network can form strong representations containing entity information accessible in a shallow way based on the target entity. We now propose a model architecture which more strongly conditions on the entity as a part of the intrinsic self-attention mechanism of the transformers.

Our approach consists of structuring input to the transformer network to use and guide the self-attention of the transformers, conditioning it on the entity. Our main mode of encoding the input, the \textbf{entity-first} method, is shown in Figure~\ref{fig:temps}. The input sequence begins with a \texttt{[START]} token, then the entity under consideration, then a \texttt{[SEP]} token. After each sentence, a \texttt{[CLS]} token is used to anchor the prediction for that sentence. In this model, the transformer can always observe the entity it should be primarily ``attending to'' from the standpoint of building representations. We also have an \textbf{entity-last} variant where the entity is primarily observed just before the classification token to condition the \texttt{[CLS]} token's self-attention accordingly.
These variants are naturally more computationally-intensive than post-conditioned models, as we need to rerun the transformer for each distinct entity we want to make a prediction for. 
\paragraph{Sentence Level vs. Document Level} As an additional variation, we can either run the transformer once per document with multiple \texttt{[CLS]} tokens (a \textbf{document-level} model as shown in Figure~\ref{fig:temps}) or specialize the prediction to a single timestep (a \textbf{sentence-level} model). In a sentence level model, we formulate each pair of entity $e$ and process step $t$ as a separate instance for our classification task. Thus, for a process with $T$ steps and $m$ entities we get $T \times m$ input sequences for fine tuning our classification task.
\begin{figure}[t!]
  \includegraphics[width=\linewidth]{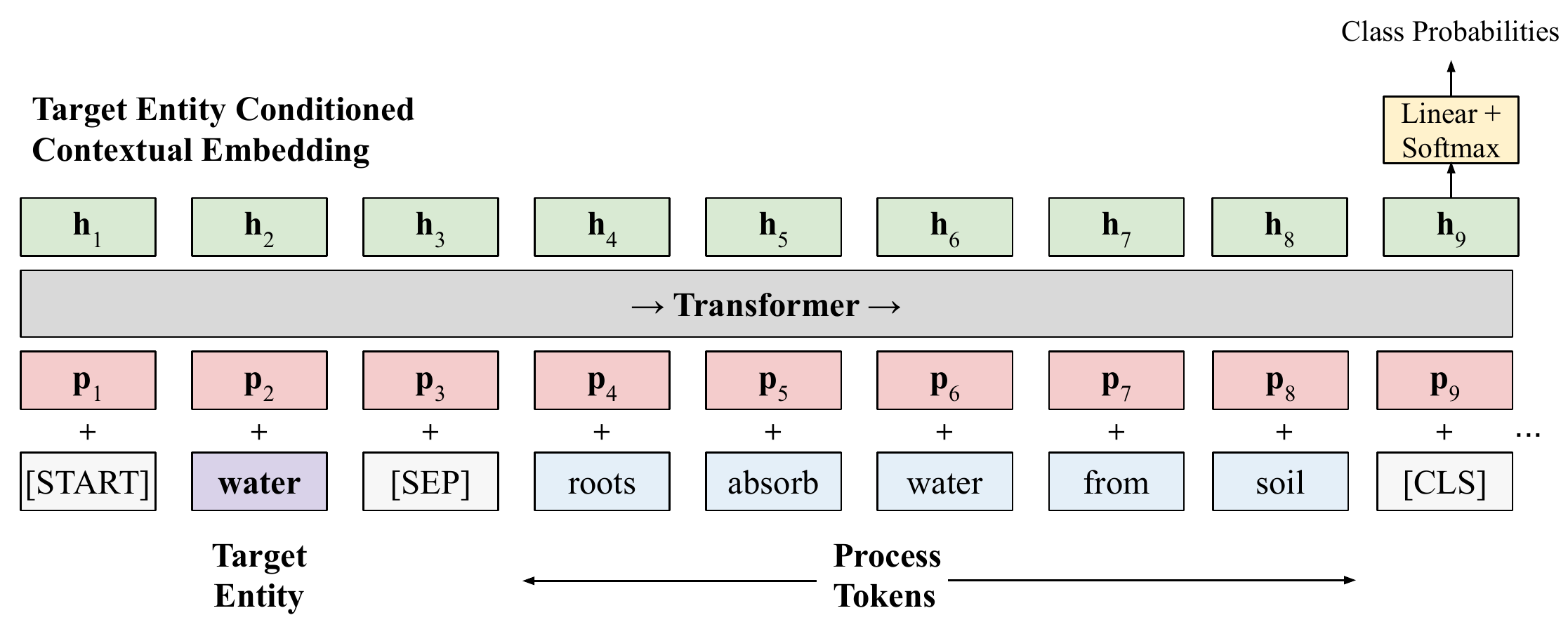}
  \centering
  \caption{Entity conditioning model for guiding self-attention: the entity-first, sentence-level input variant fed into a  left-to-right unidirectional transformer architecture. Task predictions are made at $\texttt{[CLS]}$ tokens about the entity's state after the prior sentence.}
  \label{fig:temps}
\end{figure}


\subsection{Training Details}
\label{sec:training_details}

In most experiments, we initialize the network with the weights of the standard pre-trained GPT model, then subsequently do either domain specific LM fine-tuning and supervised task specific fine-tuning.
\paragraph{Domain Specific LM fine-tuning} For some procedural domains, we have access to additional unlabeled data. To adapt the LM to capture domain intricacies, we fine-tune the transformer network on this unlabeled corpus.


\paragraph{Supervised Task Fine-Tuning} After the domain specific LM fine-tuning, we fine-tune our network parameters for the end task of entity tracking. For fine-tuning for the task, we have a labelled dataset which we denote by $\mathcal{C}$, the set of labelled pairs $(\{s_1, s_2, \dots, s_t\}, y_{et})$ for a given process. 
The input is converted according to our chosen entity conditioning procedure, then fed through the pre-trained network. 


In addition, we observed that adding the language model loss during task specific fine-tuning leads to better performance as well, possibly because it adapts the LM to our task-specific input formulation. Thus,
\begin{equation*}
\mathcal{L}_{total} = \mathcal{L}_{task} + \lambda \mathcal{L}_{lm}
\end{equation*}

\subsection{Experiments: Ingredient Detection}
We first evaluate the proposed entity conditioned self-attention model on the \textsc{Recipes} dataset to compare the performance with the post-conditioning variants. 
\subsubsection{Systems to Compare}
We use the pre-trained GPT architecture in the proposed entity conditioned framework with all its variants. BERT mainly differs in that it is bidirectional, though we also use the pre-trained \texttt{[CLS]} and \texttt{[SEP]} tokens instead of introducing new tokens in the input vocabulary and training them from scratch during fine-tuning. Owing to the lengths of the processes, all our experiments are performed on \textsc{BERT}$_{BASE}$. 

\paragraph{Neural Process Networks} 
The most significant prior work on this dataset is the work of \citet{bosselut2018simulating}. However, their data condition differs significantly from ours: they train on a large noisy training set and do not use any of the high-quality labeled data, instead treating it as dev and test data. Consequently, their model achieves low performance, roughly $56$ $F_1 $ while ours achieves $82.5$ $F_1$ (though these are not the exact same test set). Moreover, theirs underperforms the first occurrence baseline, which calls into question the value of that training data. Therefore, we do not compare to this model directly. We use the small set of human-annotated data for our probing task. Our train/dev/test split consists of $600/100/175$ recipes, respectively.


\subsubsection{Results}
Table \ref{tab:overall_recipe} compares the overall performances of our proposed models. 
Our best ET$_{GPT}$ model achieves an $F_1$ score of $82.50$. Comparing to the baselines (\emph{Majority} through \emph{First}) and post-conditioned models, we see that the early entity conditioning is critical to achieve high performance.

Although the \emph{First} model still achieves the highest CR, due to operating in a high-recall regime, we see that the ET$_{GPT}$ models all significantly outperform the post-conditioning models on this metric, indicating better modeling of these compositions. Both recall and precision are substantially increaesd compared to these baseline models. Interestingly, the ELMo-based model under-performs the first-occurrence baseline, indicating that the LSTM model is not learning much in terms of recognizing complex entity semantics grounded in long term contexts.

Comparing the four variants of structuring input in proposed architectures as discussed in Section \ref{sec:ec_models}, we observe that the \textbf{document-level, entity-first model} is the best performing variant. Given the left-to-right unidirectional transformer architecture, this model notably forms target-specific representations for all process tokens, compared to using the transformer self-attention only to extract entity specific information at the end of the process.


\begin{table}[t!]
\setlength{\tabcolsep}{3.5pt}
\small
\centering
\begin{tabular}{l|cccc|cc}
\toprule
  \bf Model & \bf P & \bf R & \bf $F_1$ & \bf Acc & \bf UR & \bf CR \\
  \midrule
   \multicolumn{7}{c}{\small Rule Based Benchmarks} \\
  \midrule
  
  Majority & - & - & - & 57.27 & - & -\\
  Exact & 84.94 & 20.25 & 32.70 & 64.39& 73.42 & 4.02 \\
  First &  65.23  & 87.17 & 74.60 & 74.65 & 84.88 & 87.79 \\
     \midrule
  \multicolumn{7}{c}{\small Post Conditioning Models} \\
  \midrule
    GPT$_{attn}$ & 63.94  & 71.72  & 67.60  & 70.63  & 54.30 & 77.04 \\
   GPT$_{concat}$ & 67.05  & 69.07 & 68.04 & 72.28 & 47.09 & 75.79 \\
    ELMo$_{token}$ & 64.96  &  76.64 & 70.32  & 72.35  & 69.14  & 78.94  \\
    ELMo$_{sent}$ & 69.09  & 72.88  & 70.90  & 74.48  & 57.05 & 77.71  \\
        \midrule
  \multicolumn{7}{c}{\small Entity-Centric Models} \\
  \midrule
  ET$_{BERT}$ & 72.49  & 80.09 & 76.10 & 78.50 & 84.30 & 78.82 \\
  ET$_{GPT}$ \textcircled{\tiny S}\textcircled{\tiny L} & 75.27 & 83.85 & 79.33 & 81.32 & 87.28 & 82.81  \\
  ET$_{GPT}$ \textcircled{\tiny S}\textcircled{\tiny F} & 76.70  & 83.98 & 80.17 & 82.26 & 88.20 & 82.69 \\
  ET$_{GPT}$ \textcircled{\tiny D}\textcircled{\tiny L} & 79.19 & 83.82 & 81.44 & 83.67 & 88.11 & 82.51 \\
  ET$_{GPT}$ \textcircled{\tiny D}\textcircled{\tiny F}  & \bf 79.85  & \bf 84.19 & \bf{81.96} & \bf 84.16 & \bf 87.91 & \bf 83.05  \\
  \bottomrule
 \end{tabular}
  \caption{Performances of different baseline models discussed in Section \ref{sec:ettn}, the ELMo baselines, and the proposed entity-centric approaches with the (D)ocument v (S)entence level variants formulated with both entity (F)irst v. (L)ater. Our ET$_{GPT}$ variants all substantially outperform the baselines.}
 \label{tab:overall_recipe}
\end{table}

\subsubsection{Ablations}
We perform ablations to evaluate the model's dependency on the context and on the target ingredient. Table \ref{tab:ing_context_recipe} shows the results for these ablations. 
\paragraph{Ingredient Specificity}  In the ``no ingredient'' baseline (w/o ing.), the model is not provided with the specific ingredient information. Table \ref{tab:ing_context_recipe} shows that while not being a strong baseline, the model achieves decent overall accuracy with the drop in UR being higher compared to CR. This indicates that there are some generic indicators (\emph{mixture}) that it can pick up to try to guess at overall ingredient presence or absence.
\begin{table}[t!]
\small
\centering
\begin{tabular}{r|cccc|cc}
\toprule
  \bf Model & \bf P & \bf R & \bf $F_1$ & \bf Acc & \bf UR & \bf CR \\
  \midrule
  \multicolumn{7}{c}{ET$_{GPT}$ \textcircled{\tiny D}\textcircled{\tiny F}} \\
  \midrule
   w/o ing. &  67.47 &  60.46  & 63.77	&	70.64 &  \textcolor{red}{35.82} & 67.97  \\
   w/ ing. &  79.85  & 84.19 & 81.96 & 84.16 & 87.91 & 83.05  \\
     \midrule
  \multicolumn{7}{c}{ET$_{GPT}$ \textcircled{\tiny S}\textcircled{\tiny F}} \\
  \midrule
  w/o context & 67.88 & 75.91 &	71.67 &	74.36 &	87.00 &	 \textcolor{red}{72.52} \\
  w/ context  & 76.70  & 83.98 & 80.17 & 82.26 & 88.20 & 82.69 \\
   \bottomrule
 \end{tabular}
 \caption{Top: we compare how much the model degrades when it conditions on no ingredient at all (w/o ing.), instead making a generic prediction. Bottom: we compare how much using previous context beyond a single sentence impacts the model.}
 \label{tab:ing_context_recipe}
\end{table}
\paragraph{Context Importance}  We compare with a ``no context'' model (w/o context) which ignore the previous context and only use the current recipe step in determining the ingredient's presence. Table \ref{tab:ing_context_recipe} shows that the such model is able to perform surprisingly well, nearly as well as the first occurrence baseline. 

This is because the model can often recognize words like verbs (for example, \emph{add}) or nouns (for example, \emph{mixture}) that indicate many ingredients are being used, and can do well without really tracking any specific entity as desired for the task.



\subsection{State Change Detection (\textsc{ProPara})}
Next, we now focus on a structured task to evaluate the performance of the entity tracking architecture in capturing the structural information in the continuous self-attention framework. For this, we use the \textsc{ProPara} dataset and evaluate our proposed model on the comprehension task. 



Figure~\ref{fig:examples}b shows an example of a short instance from the \textsc{ProPara} dataset. The task of identifying state change follows a structure satisfying the existence cycle; for example, an entity can not be created after destruction. Our prior work \cite{gupta2019tracking} proposed a structured model for the task that achieved state-of-the-art performance. 
We adapt our proposed entity tracking transformer models to this structured prediction framework, capturing creation, movement, existence (distinct from movement or creation), destruction, and non-existence.

We use the standard evaluation scheme of the \textsc{ProPara} dataset, which is framed as answering the following categories of questions: (Cat-1) \textbf{Is} \emph{e} created (destroyed, moved) in the process?,
(Cat-2) \textbf{When} (step \#) is \emph{e} created (destroyed, moved)?, (Cat-3) \textbf{Where} is \emph{e} created/destroyed/moved from/to)?


\subsubsection{Systems to Compare}

We compare our proposed models to the previous work on the \textsc{ProPara} dataset. This includes the entity specific MRC models, EntNet \cite{henaff2016tracking}, QRN \cite{seo2016query}, and KG-MRC \cite{das2018building}. Also, \citet{dalvi2018tracking} proposed two task specific models, ProLocal and ProGlobal, as baselines for the dataset. Finally, we compare against our past neural CRF entity tracking model (NCET) \cite{gupta2019tracking} which uses ELMo embeddings in a neural CRF architecture.

For the proposed GPT architecture, we use the task specific \texttt{[CLS]} token to generate tag potentials instead of class probabilities as we did previously. For BERT, we perform a similar modification as described in the previous task to utilize the pre-trained \texttt{[CLS]} token to generate tag potentials. Finally, we perform a Viterbi decoding at inference time to infer the most likely valid tag sequence.



\subsubsection{Results}

Table \ref{tab:overall_propara} compares the performance of the proposed entity tracking models on the sentence level task. Since, we are considering the classification aspect of the task, we compare our model performance for Cat-1 and Cat-2. As shown, the structured document level, entity first ET$_{GPT}$ and ET$_{BERT}$ models achieve state-of-the-art results. We observe that the major source of performance gain is attributed to the improvement in identifying the exact step(s) for the state changes (Cat-2). This shows that the model are able to better track the entities by identifying the exact step of state change (Cat-2) accurately rather than just detecting the presence of such state changes (Cat-1).
\begin{table}[t]

\small
\centering
\begin{tabular}{l|cc|cc}
\toprule
  \bf Model & \bf Cat-1 & \bf Cat-2 & \bf Ma-Avg &\bf  Mi-Avg\\
  \midrule
  \multicolumn{5}{c}{ \small Baselines} \\
  \midrule
  EntNet  & 51.62 & 18.83 & 35.22 & 37. 03\\
  QRN  & 52.37  & 15.51 & 33.94 & 35.97 \\
  ProGlobal & 62.95 & 36.39 & 49.67 & 51.13 \\
  \midrule
  \multicolumn{5}{c}{ \small Previous Work} \\
  \midrule
  KG-MRC & 62.86 & 40.00 & 51.43 & 52.69 \\
  NCET  & 70.55 & 44.57 &  57.56 & 58.99 \\
    NCET$_\text{ELMo}$ & \bf 73.68 & 47.09 & 60.38 & 61.85 \\
    \midrule
  \multicolumn{5}{c}{ \small This Work} \\
  \midrule

ET$_{GPT}$ \textcircled{\tiny D}\textcircled{\tiny F} & 73.52 & \bf 52.21 & \bf 62.87 & \bf 64.03 \\
  ET$_{BERT}$ & 73.55 & \bf 52.59 & \bf 63.07 &  \bf 64.22\\
  \bottomrule
 \end{tabular}
 \caption{Performance of the proposed models on the \textsc{ProPara} dataset. Our models outperform strong approaches from prior work across all metrics.}
 \label{tab:overall_propara}
\end{table}
This task is more highly structured and in some ways more non-local than ingredient prediction; the high performance here shows that the ET$_{GPT}$ model is able to capture document level structural information effectively. Further, the structural constraints from the CRF also aid in making better predictions. For example, in the process \textit{``higher pressure causes the sediment to heat up.  the heat causes chemical processes. the material becomes a liquid. is known as oil.''}, the \textit{material} is a by-product of the chemical process but there's no direct mention of it. However, the material ceases to exist in the next step, and because the model is able to predict this correctly, maintaining consistency results in the model finally predicting the entire state change correctly as well.

\section{Challenging Task Phenomena}
\label{sec:challenging_cases}

Based on the results in the previous section, our models clearly achieve strong performance compared to past approaches. We now revisit the challenging cases discussed in Section~\ref{sec:background} to see if our entity tracking approaches are modeling sophisticated entity phenomena as advertised. For both datasets and associated tasks, we isolate the specific set of challenging cases grounded in tracking (i) intermediate compositions formed as part of combination of entities leading to no explicit mention, and (ii) implicit events which change entities' states without explicit mention of the affects.

\subsection{Ingredient Detection}
For \textsc{Recipes}, we mainly want to investigate cases of ingredients getting re-engaged in the recipe not in a raw form but in a combined nature with other ingredients and henceforth no explicit mention. For example, \emph{eggs} in step 4 of Figure~\ref{fig:examples}a exemplifies this case. The performance in such cases is indicative of how strongly the model can track compositional entities. We also examine the performance for cases where the ingredient is referred by some other name.
 \paragraph{Intermediate Compositions} Formally, we pick the set of examples where the ground truth is a transition from $0 \rightarrow 1$ (not present to present) \emph{and} the 1 is a ``combined'' case. Table \ref{tab:intermediate_recipe} shows the model's performance on this subset of cases, of which there are $1049$ in the test set. The model achieves an accuracy of 51.1\% on these bigrams, which is relatively low given the overall model performance. In the error cases, the model defaults to the $1\rightarrow1$ pattern indicative of the \emph{First Occ} baseline. 
\begin{table}[H]
\small
\centering
\begin{tabular}{l|cccc}
\toprule
  \bf  &  $0 \rightarrow 0$ & $0 \rightarrow 1$ & $1 \rightarrow 0$ &  $1 \rightarrow 1$ \\
  \midrule
  \bf \#preds & 179 & \textbf{526} & 43 & 301 \\
\bottomrule
 \end{tabular}
 \caption{Model predictions from the document level entity first GPT model in 1049 cases of intermediate compositions. The model achieves only 51\% accuracy in these cases.}
 \label{tab:intermediate_recipe}
\end{table}

\paragraph{Hypernymy and Synonymy} We observe the model is able to capture ingredients based on their hypernyms (\textit{nuts} $\rightarrow$ \textit{pecans}, \textit{salad} $\rightarrow$ \textit{lettuce}) and rough synonymy (\textit{bourbon} $\rightarrow$ \textit{scotch}). This performance can be partially attributed to the language model pre-training. We can isolate these cases by filtering for \emph{uncombined} ingredients when there is no matching ingredient token in the step. Out of 552 such cases in the test set, the model predicts 375 correctly giving a recall of $67.9$.  This is lower than overall UR; if pre-training behaves as advertised, we expect little degradation in this case, but instead we see performance significantly below the average on uncombined ingredients.

\begin{figure*}[t!]
  \includegraphics[width=\textwidth]{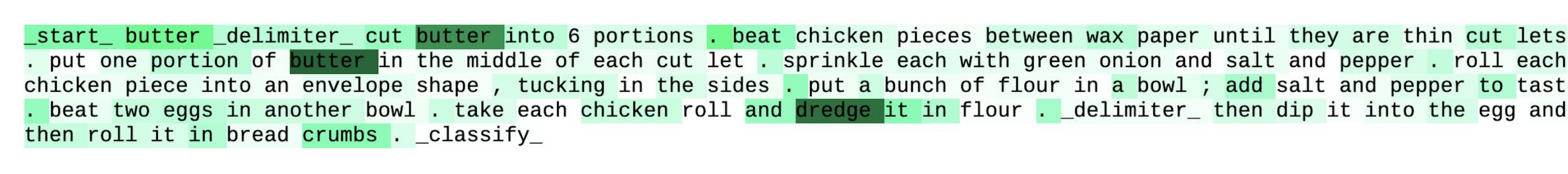}
  \centering
  \caption{Gradient of the classification loss of the gold class with respect to inputs when predicting the status of \emph{butter} in the last sentence. We follow a similar approach as \citet{jain2019attention} to compute associations. Exact matches of the entity receive high weight, as does a seemingly unrelated verb \emph{dredge}, which often indicates that the \emph{butter} has already been used and is therefore present.}
  \label{fig:gradient}
\end{figure*}

\paragraph{Impact of external data} One question we can ask of the model's capabilities is to what extent they arise from domain knowledge in the large pre-trained data. We train transformer models from scratch and additionally investigate using the large corpus of unlabeled recipes for our LM pre-training. As can be seen in Table \ref{tab:pretraining_recipe}, the incorporation of external data leads to major improvements in the overall performance. This gain is largely due to the increase in combined recall. One possible reason could be that external data leads to better understanding of verb semantics and in turn the specific ingredients forming part of the intermediate compositions. Figure~\ref{fig:gradient} shows that verbs are a critical clue the model relies on to make predictions. Performing LM fine-tuning on top of GPT also gives gains.

\subsection{State Change Detection}

\begin{table}[t!]
\small
\centering
\begin{tabular}{c|cccc|cc}
\toprule
  \bf Model & \bf P & \bf R & \bf $F_1$ & \bf Acc & \bf UR & \bf CR \\

    \midrule
  \multicolumn{7}{c}{\small No pre-training, 8 heads, 8 layers, 512 embedding size} \\
  \midrule
  No LM & 66.52   & 73.48 & 69.83  &  72.87	 & 79.20 & 71.73 \\
  20k & 72.53   & 80.32 & 76.23  & 78.59 & 79.49 & 80.58 \\
  50k & 74.40   & 81.80 & 77.92  & 80.19 & 81.90 & 81.77 \\
  \midrule
    \multicolumn{7}{c}{\small Standard GPT pre-training} \\
  \midrule
  No LM & 79.85  & 84.19 & 81.96 & 84.16 & 87.91 & 83.05  \\
  
  20k & \bf 80.14 & \bf 85.01 & \bf 82.50 & \bf 84.59 & \bf 88.83 & \bf 83.84 \\
      \bottomrule
 \end{tabular}
 \caption{Performance for using unsupervised data for LM training.}
 \label{tab:pretraining_recipe} 
\end{table}
For \textsc{ProPara}, Table \ref{tab:overall_propara} shows that the model does not significantly outperform the SOTA models in state change detection (Cat-1). However, for those correctly detected events, the transformer model outperforms the previous models for detecting the exact step of state change (Cat-2), primarily based on verb semantics. We do a finer-grained study in Table \ref{tab:separate_propara} by breaking down the performance for the three state changes: creation (C), movement (M), and destruction (D), separately. Across the three state changes, the model suffers a loss of performance in the movement cases. This is owing to the fact that the movement cases require a deeper compositional and implicit event tracking. Also, a majority of errors leading to false negatives are due to the the formation of new sub-entities which are then mentioned with other names. For example, when talking about \textit{weak acid} in \textit{``the water becomes a weak acid. the water dissolves limestone''} the \textit{weak acid} is also considered to move to the \textit{limestone}.
   
 \renewcommand{\arraystretch}{0.8}
\begin{table}[h]
\small
\centering
\begin{tabular}{l|ccc|ccc}
\toprule
  \multirow{2}{*}{\bf Model} & \multicolumn{3}{c|}{\bf \small Cat-1} & \multicolumn{3}{c}{\bf \small Cat-2} \\
  \cmidrule{2-7}
   & C & M & D & C & M & D \\
  \midrule
  ET$_{BERT}$ & 78.51 &  \textcolor{red}{61.60} & 71.50 & 76.68  &  \textcolor{red}{54.12} & 58.62\\
  ET$_{GPT}$& 79.82 &   \textcolor{red}{56.27} & 73.83 & 77.24  &  \textcolor{red}{50.82} & 56.27\\
  
  \bottomrule
 \end{tabular}
 \caption{ Results for each state change type. Performance on predicting creation and destruction are highest, partially due to the model's ability to use verb semantics for these tasks.}
 \label{tab:separate_propara}
\end{table}

\section{Analysis}

The model's performance on these challenging task cases suggests that even though it outperforms baselines, it may not be capturing deep reasoning about entities. To understand what the model actually does, we perform analysis of the model's behavior with respect to the input to understand what cues it is picking up on.

\paragraph{Gradient based Analysis} 
One way to analyze the model is to compute model gradients with respect to input features \cite{sundararajan2017axiomatic, jain2019attention}. Figure~\ref{fig:gradient} shows that in this particular example, the most important model inputs are verbs possibly associated with the entity \emph{butter}, in addition to the entity's mentions themselves. It further shows that the model learns to extract shallow clues of identifying actions exerted upon only the entity being tracked, regardless of other entities, by leveraging verb semantics.

In an ideal scenario, we would want the model to track constituent entities by translating the ``focus'' to track their newly formed compositions with other entities, often aliased by other names like \textit{mixture, blend, paste} etc. However, the low performance on such cases shown in Section \ref{sec:challenging_cases} gives further evidence that the model is not doing this.

\paragraph{Input Ablations} We can study which inputs are important more directly by explicitly removing specific certain words from the input process paragraph and evaluating the performance of the resulting input under the current model setup. We mainly did experiments to examine the importance of: (i) verbs, and (ii) other ingredients.

\begin{table}[h]
\setlength{\tabcolsep}{3.5pt}
\small
\centering
\begin{tabular}{l|c}
\toprule
  \bf Input & \bf Accuracy \\

     \midrule
  Complete Process & 84.59 \\
  \midrule 
  w/o Other Ingredients & 82.71 \\
  w/o Verbs & 79.08 \\
  w/o Verbs \& Other Ingredients & 77.79 \\

  \bottomrule
 \end{tabular}
 
 \caption{Model's performance degradation with input ablations. We see that the model's major source of performance is from verbs than compared to other ingredient's explicit mentions.}
 \label{tab:input_ablations_recipe}
\end{table}

Table \ref{tab:input_ablations_recipe} presents these ablation studies. We only observe a minor performance drop from $84.59$ to $82.71$ (accuracy) when other ingredients are removed entirely. Removing verbs dropped the performance to $79.08$ and further omitting both leads to $77.79$. This shows the model’s dependence on verb semantics over tracking the other ingredients.

\section{Conclusion}
In this paper, we examined the capabilities of transformer networks for capturing entity state semantics. First, we show that the conventional framework of using the transformer networks is not rich enough to capture entity semantics in these cases. We then propose entity-centric ways to formulate richer transformer encoding of the process paragraph, guiding the self-attention in a target entity oriented way. This approach leads to significant performance improvements, but examining model performance more deeply, we conclude that these models still do not model the intermediate compositional entities and perform well by largely relying on surface entity mentions and verb semantics.

\section*{Acknowledgments}

This work was partially supported by NSF Grant IIS-1814522 and an equipment grant from NVIDIA. The authors acknowledge the Texas Advanced Computing Center (TACC) at The University of Texas at Austin for providing HPC resources used to conduct this research. Results presented in this paper were obtained using the Chameleon testbed supported by the National Science Foundation. Thanks as well to the anonymous reviewers for their helpful comments.

\bibliography{emnlp-ijcnlp-2019}
\bibliographystyle{acl_natbib}

\appendix

\end{document}